\pdfoutput=1
\documentclass[11pt]{article}

\usepackage[a4paper,margin=1in]{geometry}
\usepackage{booktabs}
\usepackage{tabularx}
\usepackage{multirow}
\usepackage{graphicx}
\usepackage{float}
\usepackage{amsmath}
\usepackage{amssymb}
\usepackage{xcolor}
\usepackage{array}
\usepackage{subcaption}
\usepackage{url}

\title{GD-MIL: Grade-Disentangled Multiple Instance Learning
for Multimodal Biochemical Recurrence Prediction in
Prostate Cancer}

\author{
Dasari Naga Raju\\
\small \texttt{raajuuu1998@gmail.com}
}

\date{}

\begin{document}

\maketitle

\begin{abstract}
Biochemical recurrence~(BCR) after radical prostatectomy is a
clinically critical endpoint in prostate cancer, yet current
risk stratification relies almost entirely on a small number
of variables dominated by Gleason grade. Whether routine
hematoxylin and eosin~(H\&E) whole slide images~(WSIs) carry
prognostic signal \emph{beyond} grade, and whether multiple
instance learning~(MIL) can recover it, remains unsettled.
A key obstacle is that many reported pipelines select model
checkpoints on the same fold used for evaluation, artificially
inflating concordance. We construct a rigorous benchmark on
TCGA-PRAD~(487 patients, 101 BCR events) using strict
out-of-fold scoring over five-fold cross-validation repeated
across five seeds. The choice of MIL aggregator~(ABMIL,
CLAM, TransMIL, PatchGCN) has little effect~(C-index
$0.61$--$0.64$ with UNI2-h), while the feature extractor is
the dominant factor~(ResNet50 $0.566$ versus pathology
foundation models up to $0.639$). A clinical Cox model on
grade, stage, and age reaches $0.687$; no imaging-only model
significantly outperforms it~($p > 0.10$). We introduce
Grade-Disentangled MIL~(GD-MIL), a gated-attention MIL
encoder trained with a gradient-reversal grade adversary
that adversarially encourages the slide representation to be
invariant to Gleason grade before late fusion with clinical variables.
GD-MIL achieves C-index $0.704$, significantly outperforming
both the clinical baseline~($\Delta c = +0.029$,
$p = 0.0005$) and the best imaging-only model~($\Delta c =
+0.062$, $p = 0.039$), suggesting that H\&E morphology contains prognostic information
complementary to grade. A median split
on the GD-MIL risk score yields log-rank $p < 0.0001$
separation in BCR-free survival~($\sim\!20\%$ versus
$\sim\!70\%$ BCR-free survival at five years in this cohort).
\end{abstract}

\noindent\textbf{Keywords:}
prostate cancer; biochemical recurrence; multiple instance
learning; pathology foundation models; multimodal fusion;
representation disentanglement; whole slide images

\section{Introduction}
\label{sec:intro}

Prostate cancer is among the most commonly diagnosed
malignancies in men worldwide, and radical prostatectomy
is the standard curative-intent treatment for
organ-confined disease~[1]. A significant proportion of
treated patients subsequently develop biochemical
recurrence, defined as a confirmed post-operative PSA
rise above $0.2$~ng/mL~[2,3]. Because BCR typically
precedes clinical or metastatic recurrence by several
years, identifying patients at high risk at the time of
surgery is both clinically important and practically
actionable.

Current stratification relies on the Gleason grading
system~(ISUP grade groups~[4]), tumor stage, and
pre-operative PSA, combined in nomograms such as D'Amico
and CAPRA~[5,6]. While clinically useful, these tools
compress the rich morphological content of an entire
prostatectomy specimen into a handful of categorical
scores. Whether the underlying H\&E WSIs encode
prognostic information beyond what these scores already
capture remains an open and practically important
question.

Computational pathology offers a route to answering it.
Large pathology foundation models~(UNI2-h~[7],
Virchow2~[8]) combined with MIL frameworks~[9,10,11,12]
enable slide-level prediction without tile-level
annotations. For BCR specifically, however, the challenge
is harder than for grading or subtyping: the prognostic
signal in H\&E morphology is weaker and noisier, and much
of what an imaging model learns ends up as a proxy for
Gleason grade, which clinicians already record. This
leaves little independent signal for the model to exploit.
Compounding matters, many published pipelines select
model checkpoints using the evaluation fold itself,
inflating concordance estimates and making fair method
comparison impossible.

This work makes three contributions:
\begin{enumerate}
  \item A rigorous evaluation protocol on TCGA-PRAD with
        strict out-of-fold scoring across five seeds,
        eliminating checkpoint-selection leakage.
  \item A systematic benchmark showing that feature
        extractor quality is the dominant performance
        determinant, while MIL aggregator choice has
        negligible impact.
  \item Grade-Disentangled MIL~(GD-MIL), which adversarially
        discourages grade information in the learned slide
        representation via a gradient-reversal adversary
        before late fusion with clinical variables,
        achieving C-index $0.704$ and significantly
        outperforming both the clinical baseline and
        all imaging-only models.
\end{enumerate}

\section{Background}
\label{sec:background}

\subsection{Biochemical Recurrence and Prognosis}

After radical prostatectomy, BCR is defined by a sustained
post-operative PSA rise and treated as a right-censored
survival endpoint~[2,3]. The standard metric is the
censored concordance index~(C-index)~[13]: the proportion
of comparable patient pairs whose predicted risk ordering
agrees with the observed outcome ordering~($0.5$ = random
chance, $1.0$ = perfect discrimination). Reported values
for BCR prediction from H\&E WSIs typically fall in the
$0.60$--$0.70$ range. The clinically relevant question is
whether imaging \emph{adds} prognostic value beyond
clinical variables alone.

\subsection{Pathology Foundation Models}

Foundation models for computational pathology are large
vision transformers pretrained via self-supervision on
diverse WSI collections~[7,8]. UNI2-h~[7] produces
$1536$-dimensional tile features from a broad multi-cancer
collection; Virchow2~[8] produces $1280$-dimensional
features from a single large cancer centre at larger
scale. ImageNet-pretrained ResNet50~($512$-dim) serves as
a non-pathology reference.

\subsection{Multiple Instance Learning for Survival}

MIL treats a slide as a bag of $N$ tile feature vectors
with a single slide-level label. ABMIL~[9] uses gated
attention pooling to assign per-tile importance weights.
CLAM~[10] adds instance-level clustering constraints;
TransMIL~[11] introduces transformer self-attention
across tiles; PatchGCN~[12] aggregates over a spatial
graph. All are trained with the Cox partial
likelihood~[14]. We use gated-attention MIL as the
GD-MIL backbone for its interpretable per-tile weights
and natural attachment point for the grade adversary.

\subsection{Adversarial Representation Disentanglement}

Domain-adversarial training~[15] learns a representation
that is predictive for a primary task while invariant to
a nuisance variable. A gradient reversal layer~(GRL)
negates gradients from the adversary during
backpropagation, causing the encoder to suppress the
nuisance variable from its representation. We apply this to adversarially discourage grade information
from the imaging representation, encouraging grade invariance.

\section{Methods}
\label{sec:method}

\subsection{GD-MIL Architecture}
\label{sec:arch}

\begin{figure}[H]
  \centering
  \includegraphics[width=\linewidth]{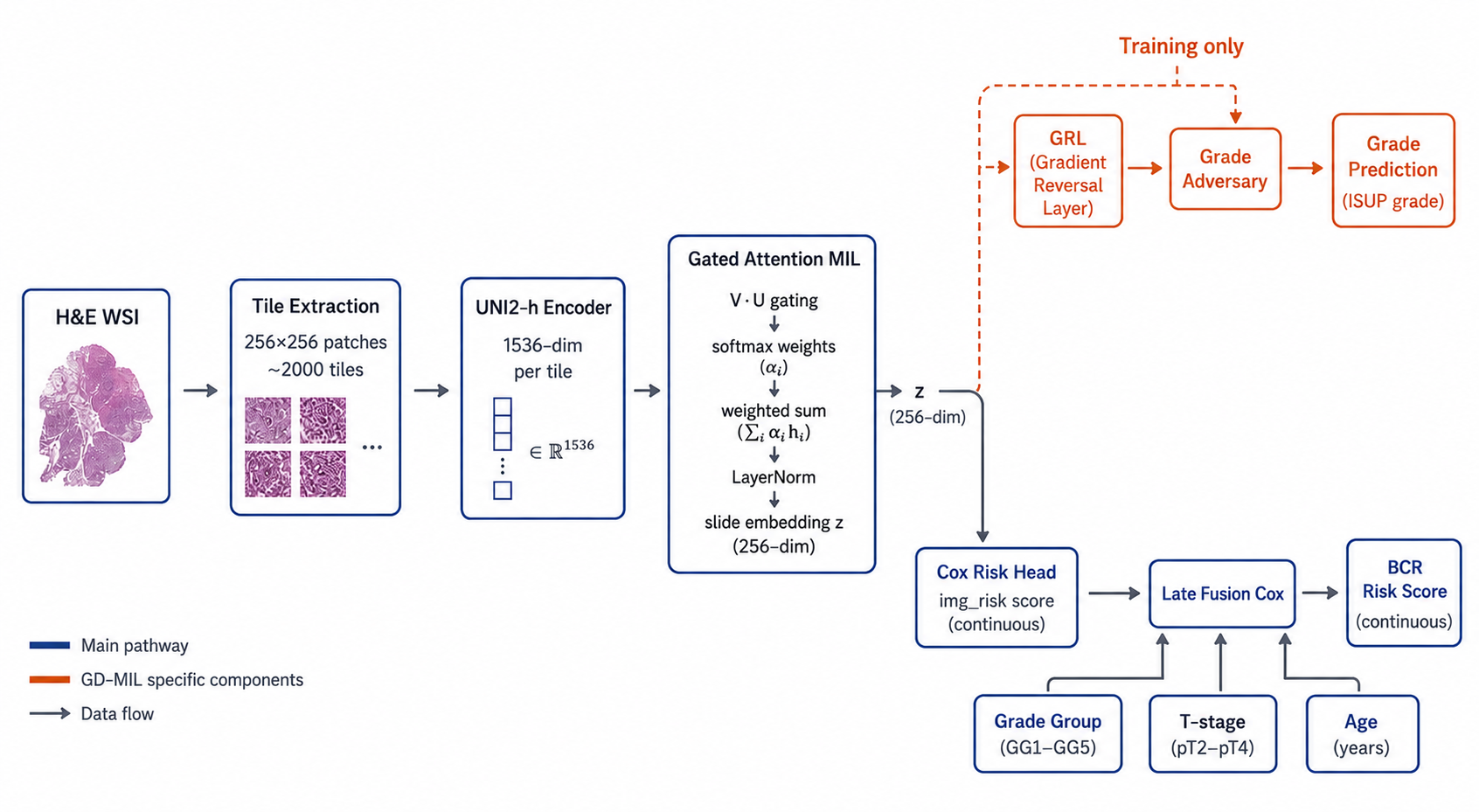}
  \caption{GD-MIL pipeline. Tile features from UNI2-h are
  aggregated by gated-attention MIL into slide
  representation~$z$. A gradient-reversal grade
  adversary~(orange, training only) encourages $z$ to be
  grade-disentangled. The imaging risk is late-fused with
  clinical variables to produce the final BCR risk score.}
  \label{fig:arch}
\end{figure}

The full pipeline is illustrated in Figure~\ref{fig:arch}.
H\&E WSI tiles~($256 \times 256$~px at $\approx 1.0$~\textmu m/px,
up to $2000$ per slide) are independently encoded by
the frozen UNI2-h model~($1536$-dim). A linear projection
maps each tile feature to $h_i \in \mathbb{R}^{256}$.
Gated attention computes normalized per-tile weights:
\begin{equation}
  a_i =
  \frac{\exp\!\big(w^\top (\tanh(V h_i) \odot \sigma(U h_i))\big)}
       {\sum_{k} \exp\!\big(w^\top (\tanh(V h_k) \odot \sigma(U h_k))\big)},
\end{equation}
where $V, U \in \mathbb{R}^{128 \times 256}$ and
$w \in \mathbb{R}^{128}$. The dual-gate mechanism~($\tanh$
relevance gate and $\sigma$ inclusion gate) suppresses
uninformative tiles more effectively than a single gate.
The attention-weighted sum $\sum_i a_i h_i$ is
layer-normalized to give slide representation
$z \in \mathbb{R}^{256}$.

The architecture has two branches. The \textbf{main branch}
produces imaging risk $r_{\text{img}} = \phi(z)$ via a
linear Cox risk head. The \textbf{adversarial branch},
active only during training, attaches a two-layer MLP
grade predictor $\psi$ through a GRL:
\begin{equation}
  \hat{g} = \psi\!\big(\mathcal{R}_\lambda(z)\big),
  \qquad
  \frac{\partial \mathcal{R}_\lambda}{\partial z} = -\lambda I.
\end{equation}
The reversed gradient encourages the encoder to suppress
grade-predictive information while retaining BCR-prognostic
signal. The combined training objective is:
\begin{equation}
  \mathcal{L}
  = \mathcal{L}_{\text{cox}}(r_{\text{img}})
  + \lambda \, \big\| \hat{g} - g \big\|_2^2,
  \label{eq:gdmil-loss}
\end{equation}
where $g$ is the standardized ISUP grade and $\lambda =
0.5$. The
grade-disentangled risk $r_{\text{img}}$ is concatenated
with clinical variables~(grade group, T-stage, age) and
a late-fusion Cox model produces the final GD-MIL risk
score, contributing complementary information to the
clinical baseline.

\subsection{Evaluation Protocol}
\label{sec:eval}

We use five-fold stratified cross-validation~(stratified
by BCR event status). Within each fold, a $15\%$ inner
validation split is reserved exclusively for early
stopping; the outer test fold is never used for model
selection. Each patient receives a single out-of-fold~(OOF)
risk prediction, and the C-index is computed over all
$487$ OOF predictions. The procedure is repeated across
five independent random seeds; we report mean and standard
deviation across seeds. Pairwise significance is assessed
by paired bootstrap tests with $2000$ patient-level
resamples, and $95\%$ confidence intervals are computed
by the same bootstrap.

\subsection{Baselines}

The \textbf{clinical baseline} is an $\ell_2$-penalized
Cox model~[14] on ISUP grade group~(1--5), ordinal
T-stage, and age at surgery. \textbf{Imaging baselines}
are MIL models~(ABMIL, CLAM, TransMIL, PatchGCN) trained
on UNI2-h, Virchow2, and ResNet50 features under the
same evaluation protocol.

\subsection{Cox Partial Likelihood}
\label{sec:cox}

Given risk scores $r_i$, event indicators $\delta_i$,
and observed times $t_i$:
\begin{equation}
  \mathcal{L}_{\text{cox}}
  = -\frac{1}{\sum_i \delta_i}
    \sum_{i:\,\delta_i = 1}
    \left(
      r_i - \log \sum_{j:\,t_j \ge t_i} e^{r_j}
    \right).
\end{equation}

\subsection{Data and Feature Extraction}

\textbf{TCGA-PRAD.} We used $487$ patients from the TCGA
Prostate Adenocarcinoma cohort~[16]~($101$ BCR events,
$20.7\%$; median follow-up $559$ days) with matched H\&E
WSIs, ISUP grade group, ordinal T-stage, and age.
Missing values were median-imputed. Tissue was segmented
by an HSV filter and tiled at $\approx 1.0$~\textmu m/px~($256
\times 256$~px; up to $2000$ tiles per slide). UNI2-h~($d
= 1536$), Virchow2~($d = 1280$), and ResNet50~($d = 512$)
features were extracted without stain normalization or
augmentation. Tile coordinates were saved for attention
visualization.

\subsection{Training Details}

All models used Adam~[17]~(lr $3 \times 10^{-4}$, weight
decay $10^{-5}$), gradient clipping at norm $1.0$, and
early stopping on the inner validation fold. GD-MIL used
$\lambda = 0.5$. Cox and fusion models used $\ell_2$
penalization of $0.1$. All fusion steps were fitted
exclusively on OOF imaging risk scores.

\section{Results}
\label{sec:results}

\subsection{Benchmark: Feature Extractor Quality Dominates}

\begin{figure}[H]
  \centering
  \includegraphics[width=0.92\linewidth]{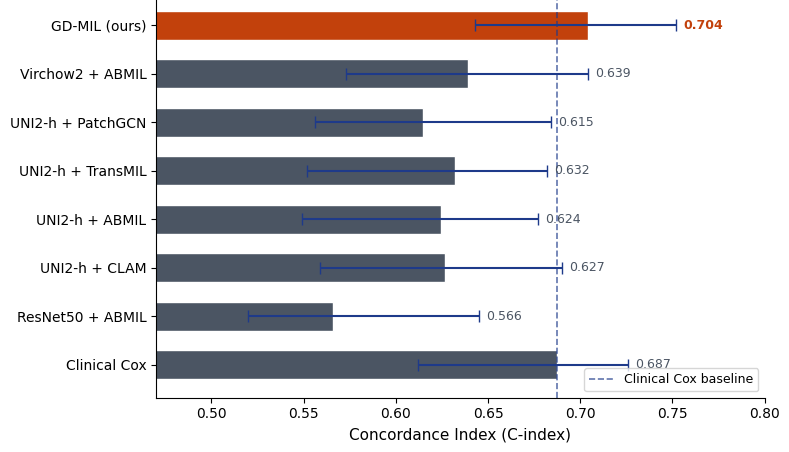}
  \caption{C-index comparison on TCGA-PRAD. Error bars:
  95\% bootstrap CI~(2000 resamples). Dashed line: clinical
  Cox baseline~($0.687$). GD-MIL~(orange) is the only
  method whose full CI lies above the clinical baseline.}
  \label{fig:cindex}
\end{figure}

Table~\ref{tab:results} and Figure~\ref{fig:cindex} report
all methods under the evaluation protocol. The results
reveal a clear performance hierarchy. ResNet50~+~ABMIL
reaches C-index $0.566$, barely above random chance,
while all UNI2-h aggregators cluster tightly between
$0.615$ and $0.632$ regardless of aggregator choice.
This $0.017$-point range across four MIL architectures
falls within seed-to-seed variability~(std up to $0.036$),
suggesting that the MIL aggregator is not a meaningful
performance lever once high-quality features are in place.
By contrast, the transition from ResNet50 to UNI2-h
yields a $\approx 0.06$ C-index gain, suggesting
feature extractor quality as the dominant factor.
Virchow2~+~ABMIL at $0.639$ is consistent with its
larger pretraining scale.

The clinical Cox model at $0.687$ forms a ceiling for all
imaging-only approaches. No imaging-only model
significantly outperforms it~(Table~\ref{tab:significance},
$p = 0.11$--$0.31$), indicating that H\&E features alone do not substitute
for clinical risk variables on this cohort. GD-MIL at $0.704$
is the only method whose full bootstrap CI~$[0.643,\
0.752]$ lies entirely above the clinical baseline
~$[0.612,\ 0.726]$, and significantly outperforms both
the clinical Cox~($\Delta c = +0.029$, $p = 0.0005$) and
the best imaging-only model~(Virchow2~+~ABMIL,
$\Delta c = +0.062$, $p = 0.039$). The low cross-seed
variance~(std $0.003$) is consistent across seeds, suggesting stability of the result. The $+0.085$ gain over plain ABMIL~+~UNI2-h
~($p = 0.003$) further supports the interpretation that
the grade-adversarial design, not merely the use of a
foundation model, contributes to the improvement.

\begin{table}[H]
\caption{Results on TCGA-PRAD~(487 patients, 101 BCR
events). C-index: mean $\pm$ std across five seeds,
scored strictly out-of-fold. 95\% CI from 2000 bootstrap
resamples. $^\dagger$GD-MIL fuses grade-disentangled
imaging risk with clinical variables via a late-fusion
Cox model.}
\label{tab:results}
\centering
\small
\begin{tabular}{llcc}
\toprule
\textbf{Category} & \textbf{Method} &
\textbf{C-index} & \textbf{95\% CI} \\
\midrule
Clinical &
  Cox (grade, T, age) &
  $0.687 \pm 0.005$ & $[0.612,\ 0.726]$ \\
\midrule
\multirow{6}{*}{Imaging}
  & ResNet50 + ABMIL   & $0.566 \pm 0.024$ & $[0.520,\ 0.645]$ \\
  & UNI2-h + CLAM      & $0.627 \pm 0.022$ & $[0.559,\ 0.690]$ \\
  & UNI2-h + ABMIL     & $0.624 \pm 0.026$ & $[0.549,\ 0.677]$ \\
  & UNI2-h + TransMIL  & $0.632 \pm 0.022$ & $[0.552,\ 0.682]$ \\
  & UNI2-h + PatchGCN  & $0.615 \pm 0.036$ & $[0.556,\ 0.684]$ \\
  & Virchow2 + ABMIL   & $0.639 \pm 0.007$ & $[0.573,\ 0.704]$ \\
\midrule
Multimodal &
  \textbf{GD-MIL (ours)}$^\dagger$ &
  $\mathbf{0.704 \pm 0.003}$ & $\mathbf{[0.643,\ 0.752]}$ \\
\bottomrule
\end{tabular}
\end{table}

\begin{table}[H]
\caption{Paired bootstrap significance tests~(2000
resamples). $\Delta c$: C-index of row minus comparator.
$^\star p < 0.05$;\ $^\dagger p < 0.01$;\
$^\ddagger p < 0.001$.}
\label{tab:significance}
\centering
\small
\begin{tabular}{llccc}
\toprule
\textbf{Method} & \textbf{Comparator} &
$\boldsymbol{\Delta c}$ & \textbf{95\% CI} &
\textbf{\textit{p}} \\
\midrule
GD-MIL (ours) & Clinical Cox &
  $+0.029$ & $[+0.015,\ +0.046]$ & $0.0005^{\ddagger}$ \\
GD-MIL (ours) & ABMIL / UNI2-h &
  $+0.085$ & $[+0.026,\ +0.144]$ & $0.003^{\dagger}$ \\
GD-MIL (ours) & ABMIL / Virchow2 &
  $+0.062$ & $[+0.004,\ +0.118]$ & $0.039^{\star}$ \\
\midrule
ABMIL / Virchow2 & Clinical Cox &
  $-0.033$ & $[-0.097,\ +0.031]$ & $0.310$ \\
ABMIL / UNI2-h & Clinical Cox &
  $-0.056$ & $[-0.124,\ +0.011]$ & $0.112$ \\
\bottomrule
\end{tabular}
\end{table}

\subsection{Risk Stratification}

\begin{figure}[H]
  \centering
  \includegraphics[width=0.85\linewidth]{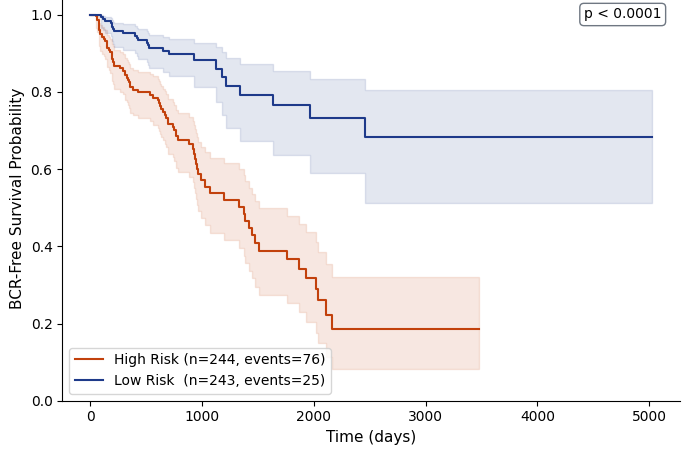}
  \caption{Kaplan-Meier BCR-free survival by median
  GD-MIL risk score. High-risk: $n = 244$, 76 events.
  Low-risk: $n = 243$, 25 events. Log-rank $p < 0.0001$.
  Shaded bands: 95\% CI.}
  \label{fig:km}
\end{figure}

To assess clinical utility beyond concordance,
Figure~\ref{fig:km} shows Kaplan-Meier BCR-free survival
curves stratified by median GD-MIL risk score. The
high-risk group~($n = 244$, $76$ events, $31.1\%$) and
low-risk group~($n = 243$, $25$ events, $10.3\%$) separate
immediately after surgery and remain well-separated
throughout five years of follow-up with no crossover.
The three-fold difference in event rate~($31.1\%$ versus
$10.3\%$) is consistent with the separation reflecting
differences in recurrence burden rather than differential
censoring, though this cannot be formally excluded on a
single cohort.
At five years, the high-risk group reaches $\sim\!20\%$
BCR-free survival versus $\sim\!70\%$ in the low-risk group,
a $50$-percentage-point gap~(log-rank $p < 0.0001$).
This separation could directly inform post-operative
management: high-risk patients are strong candidates for
adjuvant radiotherapy or intensified PSA monitoring,
while low-risk patients may be safely observed.

\subsection{Attention Analysis}

\begin{figure}[H]
  \centering
  \includegraphics[width=\linewidth]{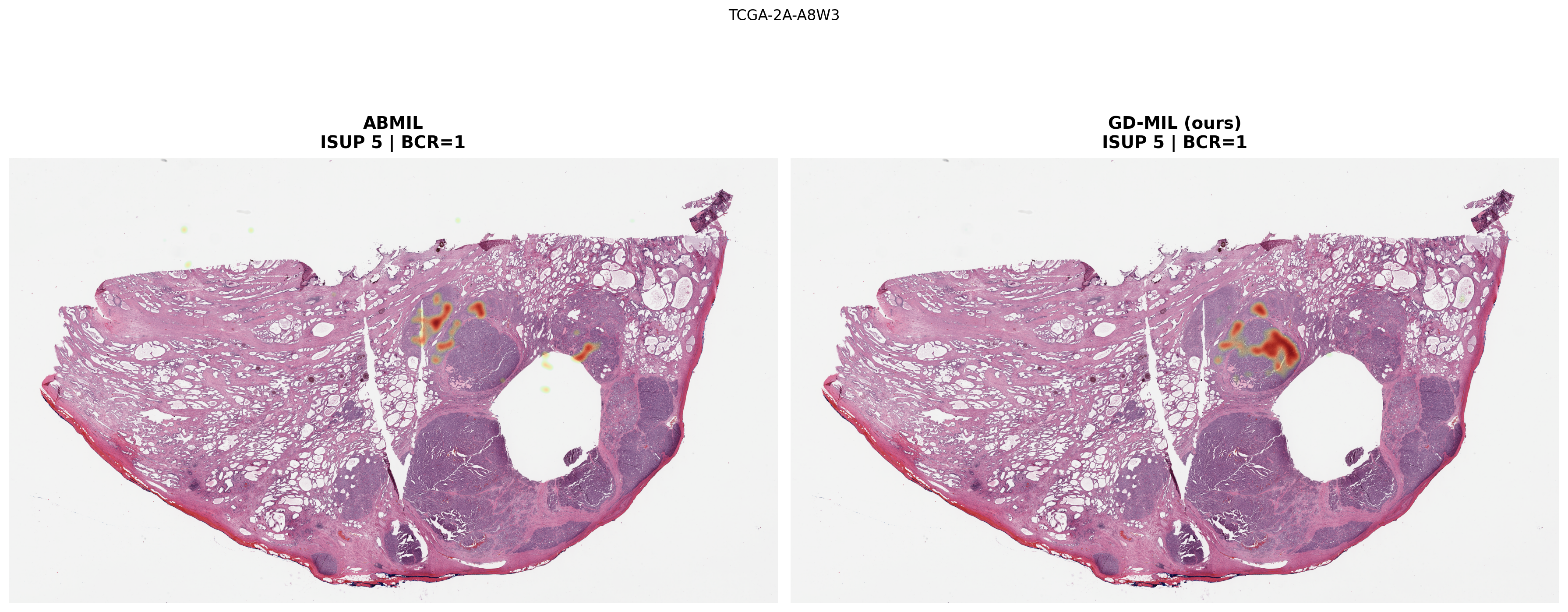}
  \caption{Attention maps, tighter colormap scale
  (TCGA-2A-A8W3, ISUP~5, BCR$=$1). Left: ABMIL attends
  to a visually prominent region in the left-centre.
  Right: GD-MIL redirects to a distinct morphological
  region in the right-centre, qualitatively consistent
  with grade-disentangled attention.}
  \label{fig:attn_a}
\end{figure}

\begin{figure}[H]
  \centering
  \includegraphics[width=\linewidth]{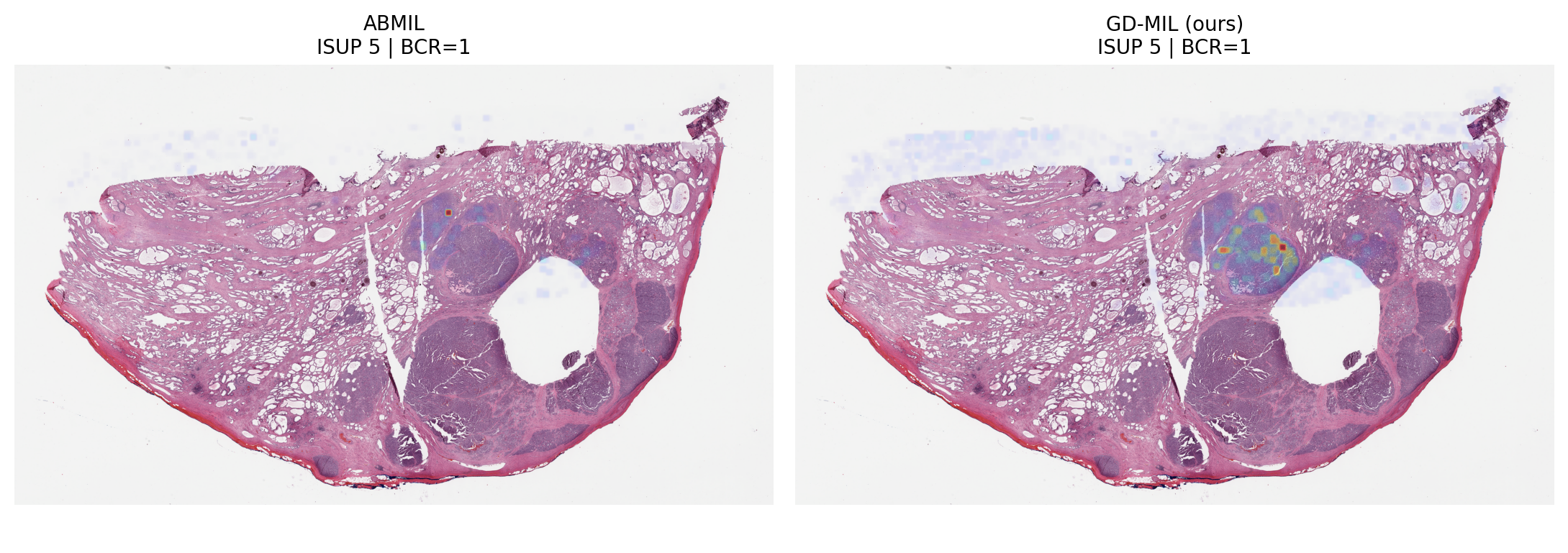}
  \caption{Same case, broader colormap scale. ABMIL
  produces near-zero attention across the slide. GD-MIL
  retains a sharp focal activation in the right-centre
  region, demonstrating more decisive attention weights.
  Tile colour: cool = low attention, warm/orange = high.}
  \label{fig:attn_b}
\end{figure}

Figures~\ref{fig:attn_a} and~\ref{fig:attn_b} compare
attention maps from ABMIL and GD-MIL on a representative
high-grade BCR-positive case~(TCGA-2A-A8W3, ISUP~5,
BCR$=$1). At the tighter colormap scale~(Figure~\ref{fig:attn_a}),
ABMIL concentrates on a visually prominent tumour-containing
region in the left-centre of the slide, consistent with a
region that a grade-correlated model might preferentially
attend to.
GD-MIL redirects attention to a distinct morphological
region in the right-centre, qualitatively consistent
with the grade-adversarial objective — though
interpretation of a single case is necessarily illustrative. At the broader scale~(Figure~\ref{fig:attn_b}),
ABMIL produces near-zero attention weights across the
entire slide, suggesting weaker localization of
discriminative regions. GD-MIL retains a sharp focal
activation in the same right-centre region, indicating
more spatially concentrated attention weights.
Rigorous interpretation would require pathologist-annotated
regions of interest.

\section{Discussion}
\label{sec:discussion}

GD-MIL significantly outperforms both the clinical
baseline~($p = 0.0005$) and the best imaging-only
model~($p = 0.039$). Because the clinical model already includes Gleason
grade, this result is consistent with H\&E WSIs
containing morphological prognostic information
complementary to grade, though external validation
would be needed to confirm generalisability. The modest effect size~($\Delta c = +0.029$) is not
surprising given the inherent noise of BCR as an endpoint,
the relatively small number of events~($101$ of $487$), and
the single-cohort design. What matters more is that the
improvement is highly consistent across seeds~(std
$0.003$) and survives rigorous paired significance testing.

Beyond GD-MIL itself, the benchmark offers a
practical finding for the field. The $\approx 0.06$
C-index gap between ResNet50 and UNI2-h substantially
outweighs the $0.017$ spread across all four UNI2-h
aggregators, indicating that feature extractor quality
matters far more than aggregator choice. For practitioners
building BCR prediction pipelines, this suggests that
investment in high-quality tile features will yield
more return than architectural complexity.

The $+0.085$ gain over plain ABMIL~+~UNI2-h~($p =
0.003$) supports the view that grade disentanglement,
rather than the foundation model alone, drives this
improvement. The Kaplan-Meier curves put a clinical face
on these numbers: a $50$-percentage-point BCR-free
survival gap at five years~(log-rank $p < 0.0001$)
between median-split risk groups, potentially informative
for decisions about adjuvant radiotherapy and follow-up
intensity.

\textbf{Limitations.} Results are from TCGA-PRAD only;
performance may not transfer across institutions due to
differences in slide preparation, scanning hardware, and
patient population. The clinical feature set excludes
pre-operative PSA and surgical margin status, both of
which carry known prognostic value. The tile cap of
$2000$ may discard informative tissue on large specimens.
Grade disentanglement is enforced during training but not
formally verified at inference; measuring residual grade
information via mutual information or probing classifiers
would provide stronger validation.

\section{Conclusion}
\label{sec:conclusion}

We introduced GD-MIL, a grade-disentangled
attention-MIL framework that adversarially discourages
Gleason grade information from the learned slide
representation before late fusion with clinical variables.
On TCGA-PRAD~(487 patients, 101 BCR events), GD-MIL
achieves C-index $0.704$, significantly outperforming both
the clinical baseline~($\Delta c = +0.029$, $p = 0.0005$)
and all imaging-only models~($\Delta c = +0.062$,
$p = 0.039$), consistent with H\&E morphology encoding
prognostic information beyond what grade alone captures.
The accompanying benchmark shows that feature extractor
quality, not aggregator choice, is the dominant
performance lever across MIL methods. A median risk split
yields a $50$-percentage-point BCR-free survival
gap~(log-rank $p < 0.0001$) at five years, suggesting
clinical utility warranting external validation.

\section*{Code Availability}

The evaluation protocol, all baselines, and GD-MIL are
publicly available at
\url{https://github.com/raajuuu1998/gd-mil-bcr}.



\begin{thebibliography}{99}

\bibitem[1]{ref-sung2021}
H.~Sung et al.
\newblock Global Cancer Statistics 2020.
\newblock {\em CA: A Cancer Journal for Clinicians},
71(3):209--249, 2021.

\bibitem[2]{ref-pound1999}
C.R.~Pound et al.
\newblock Natural History of Progression After PSA Elevation
Following Radical Prostatectomy.
\newblock {\em JAMA}, 281(17):1591--1597, 1999.

\bibitem[3]{ref-freedland2005}
S.J.~Freedland et al.
\newblock Risk of Prostate Cancer-Specific Mortality Following
Biochemical Recurrence After Radical Prostatectomy.
\newblock {\em JAMA}, 294(4):433--439, 2005.

\bibitem[4]{ref-epstein2016}
J.I.~Epstein et al.
\newblock A Contemporary Prostate Cancer Grading System.
\newblock {\em European Urology}, 69(3):428--435, 2016.

\bibitem[5]{ref-damico1998}
A.V.~D'Amico et al.
\newblock Biochemical Outcome After Radical Prostatectomy.
\newblock {\em JAMA}, 280(11):969--974, 1998.

\bibitem[6]{ref-cooperberg2005}
M.R.~Cooperberg et al.
\newblock The UCSF Cancer of the Prostate Risk Assessment Score.
\newblock {\em The Journal of Urology}, 173(6):1938--1942, 2005.

\bibitem[7]{ref-chen2024uni}
R.J.~Chen et al.
\newblock Towards a General-Purpose Foundation Model for
Computational Pathology.
\newblock {\em Nature Medicine}, 30:850--862, 2024.

\bibitem[8]{ref-vorontsov2024}
E.~Vorontsov et al.
\newblock A Foundation Model for Clinical-Grade Computational
Pathology and Rare Cancers Detection.
\newblock {\em Nature Medicine}, 30:2924--2935, 2024.

\bibitem[9]{ref-ilse2018}
M.~Ilse, J.M.~Tomczak, and M.~Welling.
\newblock Attention-Based Deep Multiple Instance Learning.
\newblock In {\em ICML}, pages 2127--2136, 2018.

\bibitem[10]{ref-lu2021}
M.Y.~Lu et al.
\newblock Data-Efficient and Weakly Supervised Computational
Pathology on Whole-Slide Images.
\newblock {\em Nature Biomedical Engineering}, 5(6):555--570,
2021.

\bibitem[11]{ref-shao2021}
Z.~Shao et al.
\newblock TransMIL: Transformer Based Correlated Multiple
Instance Learning for Whole Slide Image Classification.
\newblock In {\em NeurIPS}, volume~34, 2021.

\bibitem[12]{ref-chen2021patchgcn}
R.J.~Chen et al.
\newblock Whole Slide Images Are 2D Point Clouds.
\newblock In {\em MICCAI}, pages 339--349, 2021.

\bibitem[13]{ref-harrell1982}
F.E.~Harrell et al.
\newblock Evaluating the Yield of Medical Tests.
\newblock {\em JAMA}, 247(18):2543--2546, 1982.

\bibitem[14]{ref-cox1972}
D.R.~Cox.
\newblock Regression Models and Life-Tables.
\newblock {\em JRSS-B}, 34(2):187--202, 1972.

\bibitem[15]{ref-ganin2016}
Y.~Ganin et al.
\newblock Domain-Adversarial Training of Neural Networks.
\newblock {\em JMLR}, 17(59):1--35, 2016.

\bibitem[16]{ref-tcga2015prad}
The Cancer Genome Atlas Research Network.
\newblock The Molecular Taxonomy of Primary Prostate Cancer.
\newblock {\em Cell}, 163(4):1011--1025, 2015.

\bibitem[17]{ref-adam}
D.P.~Kingma and J.~Ba.
\newblock Adam: A Method for Stochastic Optimization.
\newblock In {\em ICLR}, 2015.

\bibitem[18]{ref-leopard2024}
LEOPARD Challenge Organizers.
\newblock LEOPARD: LEarning biOchemical Prostate cAncer
Recurrence from histopathology sliDes.
\newblock MICCAI Challenge, 2024.

\end{thebibliography}
\end{document}